**A Century Long Commitment to Assessing Artificial Intelligence and Its Impact on Society**[1]
Barbara J. Grosz, Harvard University, Inaugural Chair of the AI100 Standing Committee
Peter Stone, University of Texas at Austin, Chair of the Inaugural AI100 Study Panel

The Stanford One Hundred Year Study on Artificial Intelligence, a project that launched in December 2014, is designed to be a century-long periodic assessment of the field of Artificial Intelligence (AI) and its influences on people, their communities, and society. Colloquially referred to as "AI100", the project issued its first report in September 2016. A Standing Committee works with the Stanford Faculty Director of AI100 in overseeing the project and designing its activities. A little more than two years after the first report appeared, we reflect on the decisions made in shaping it, the process that produced it, its major conclusions, and reactions subsequent to its release.

The inaugural AI100 report [4], which is titled "Artificial Intelligence and Life in 2030," examines eight domains of human activity in which AI technologies are already starting to affect urban life. In scope, it encompasses domains with emerging products enabled by AI methods and ones raising concerns about technological impact generated by potential AI-enabled systems. The Study Panel members who authored the report and the AI100 Standing Committee, which is the body that directs the AI100 project, intend for it to act as a catalyst, spurring conversations on how we as a society might shape and share the potentially powerful technologies that AI could enable. In addition to influencing researchers and guiding decisions in industry and governments, the report aims to provide the general public with a scientifically and technologically accurate portrayal of the current state of AI and its potential. It aspires to replace conceptions rooted in science fiction books and movies with a realistic foundation for these deliberations.

The report focuses on AI research and "specialized AI technologies" — methods developed for and tailored to particular applications — that are increasingly prevalent in daily activities rather than deliberating about generalized intelligence, which is often mentioned in the media and is much further from realization. It anticipates that AI-enabled systems have great potential to improve daily life and have positive impact on economies worldwide, but also will create profound societal and ethical challenges. It thus argues that deliberations involving the broadest possible spectrum of expertise about AI technologies and the design, ethical, and policy challenges they raise should begin now to ensure that the benefits of AI are broadly shared as well as that systems are safe, reliable, and trustworthy.

In the remainder of this column, we provide background on AI100 and the framing of its first report and then discuss some of the findings. Along the way, we address several questions posed to us during the years since the report appeared, and catalog some of its uses.

**Influences and Origins of the One Hundred Year Study**

The impetus for the AI100 study came from the many positive responses to a 2008-2009 AAAI Presidential Panel on Long-Term AI Futures that was commissioned by then AAAI President Eric Horvitz (Microsoft Research) and co-chaired by him and Bart Selman (Cornell University). Intending a largely field-internal reflection on the state of AI, Horvitz charged the panel with exploring "the potential long-term societal influences of AI advances." In particular, he asked them to consider such issues as AI successes and the societal opportunities and challenges they

---

raised; the socioeconomic, ethical and legal issues raised by AI technologies; proactive steps those in the field could take to enhance long-term societal outcomes; and the kinds of policies and guidelines needed for autonomous systems.  The findings of this panel (which may be found at http://www.aaai.org/Organization/presidential-panel.php) and reactions to it led Horvitz to design AI100, a long-horizon study of influences of AI advances on people and society. This study is to undertake periodic studies of developments, trends, futures, and potential disruptions associated with developments in machine intelligence, and to formulate assessments, recommendations and guidance on proactive efforts. The new project was to be balanced in its inward (within field) and outward (other disciplines and society at large) looking faces. The long-term nature of the project is its most novel aspect; it is intended to periodically (typically every 5 years) over the course of at least 100 years assemble a study panel to reassess the state of AI and its impact on society.  A "framing memo" (https://ai100.stanford.edu/reflections-and-framing) lays out Horvitz's aspirations for this project along with the reasons for situating it at Stanford University.

**Shaping the first step of a 100-year project:**

*Assembling a Study Panel:* As the AI100 project was launched in December 2014, it was anticipated that several years would be available for shaping the project, engaging people with expertise across the social sciences and humanities as well as AI in this effort, identifying a focal topic, and recruiting a study panel.  Within a few months, however, it became apparent that AI was entering daily life and garnering intense public interest and scrutiny at a rate that did not allow such a leisurely pace. The Standing Committee defined a compressed schedule, and it recruited Peter Stone, The University of Texas at Austin (co-author of this column) as the chair of the report's Study Panel.  Together they assembled a 17-member Study Panel comprising experts in AI from academia, corporate laboratories and industry, and AI-savvy scholars in law, political science, policy, and economics. Although their goal was a panel diverse in specialty and expertise, geographic region, gender, and career stages, the shortened time frame led to the Study Panel being less geographically and field diverse than ideal, a point noted by some report readers. In recognition of these shortcomings, as it considers the design of future studies, the Steering Committee, which has increased its membership to include more representation from the social sciences and humanities, is currently designing a more inclusive planning and reporting process.

*Designing the Charge:* The Standing Committee considered a variety of possible themes and scopes for the inaugural AI100 report, ranging from a general survey of the status of research and applications in subfields to an in-depth examination of a particular technology, such as machine learning or natural language processing, or an application area, such as healthcare or transportation. Its final choice reflects a desire to ground the report's assessments in a context that would bring to the fore societal settings and a broad array of technological developments. The focus on "AI and Life in 2030" arose from a recognition of the central role cities have played throughout most of human history as well as of cities as a venue in which many AI technologies are likely to come together in the lives of individuals and communities. The further focus on North American cities followed from a recognition that within the short time frame of the Panel's work, it was not possible to adequately consider the great variability of urban settings and cultures around the world. Although the Standing Committee expects that the projections, assessments, and proactive guidance stemming from the study will have broader global relevance, it intends for future studies to have greater international involvement and scope.

The charge sent to the Study Panel asked it to identify possible advances in AI over 15 years and their potential influences on daily life in urban settings (with a focus on North American cities), to specify scientific, engineering, and policy and legal efforts needed to realize these developments, to consider actions needed to shape outcomes for societal good and to deliberate on the design, ethical, and policy challenges the developments raise. It further stipulated that the Study Panel should ground their examination of AI technologies in a context that highlights interdependencies and interactions among AI subfields, these technologies and their potential influences on a wide variety of activities.[2]

*Creating the First Report:* In the absence of precedent and with a short time horizon, the Study Panel engaged in a sequence of virtually convened brainstorming sessions. During these sessions, they successively refined the topics to consider in their report, with the aim of identifying domains, or economic sectors, in which AI seemed most likely to have impact within urban settings over the coming 15 years. Then during a full-day intensive writing session during an in-person meeting in February at the 2016 AAAI conference, they drafted report sections. They iteratively revised these section drafts with the goal of producing a report that was accessible to the general public and conveyed the Study Panel's key messages. At a final in-person meeting in July at the 2017 IJCAI conference, the Study Panel identified the main messages of the report, which appear as callouts in the margins of the report.[3]

**The Report**

The report aims to address several different audiences, ranging from the general public to AI researchers and practitioners, and thus to both be accessible and to provide depth. As a result, the report has a 3-part hierarchical structure: an executive summary, a more expansive 5-page overview, which summarizes the core of the report, and the core with further details. The core examines eight "domains" of typical urban settings on which AI is likely to have impact over the coming years: transportation, home and service robots, healthcare, education, public safety and security, low-resource communities, employment and workplace, and entertainment. The authors deliberately did not give much weight to positions they considered excessively optimistic or pessimistic, despite the prevalence of such positions in the popular press, because they intended the report to provide a balanced and sober assessment by the people at the heart of technological developments in AI.

For each domain the Study Panel investigated, the report looks back 15 years to summarize the AI-enabled changes that have already occurred, and then projects forward 15 years. It identifies the availability of large amounts of data, including speech and geospatial data, as well as cloud computing resources and progress in hardware technology for sensing and perception, as contributing to recent advances in AI research as well as to the success of deployed AI-enabled systems. Advances in machine learning, fueled in part by these resources, as well as the development of "deep" artificial neural nets, have played a key role in enabling these achievements. The goal of the Study Panel's forward-looking assessment, which we summarize briefly, was to call attention to the opportunities the Study Panel saw for AI technologies to improve societal conditions, to lower the barriers to realizing this potential, and to address the realistic risks it saw in applying AI technologies in the domains it studied.

---

2        As the report notes, one consequence of the decision to focus the first study on urban life was that military applications of AI were outside its scope. The Standing Committee recognizes the importance of monitoring and deliberation about the implications of AI advances for military applications and expects these issues to be considered by future study panels.
3        AAAI, the Association for the Advancement of AI, is a major, and the oldest, scientific society for the field. IJCAI, the International Joint Conferences on Artificial Intelligence, is a major, and the oldest, international AI conference.

The projected time horizons for AI-enabled systems to enter daily life vary across these domains, as do the opportunities for transforming life and the challenges posed in each domain. Furthermore, the challenges identified by the Study Panel ranged across the full spectrum of computer science from hardware to human-computer interaction. For instance, improvements in safe and reliable hardware were determined to be essential for progress in transportation and home-service robots. Autonomous transportation, which the report projects, may "soon be commonplace", is among the currently most visible AI applications; in addition to changing individuals' driving needs, it is expected to affect transportation infrastructure, urban organization, and jobs.  Experience with home-service robots illustrates the key role of hardware: Although robotic vacuum cleaners have been in home use for many years, technical constraints and the high costs of reliable mechanical devices has limited commercial opportunities to narrowly defined applications; the report projects they will do so for the foreseeable future. For healthcare, the challenges highlighted in the report include developing mechanisms for sharing data, removing policy, regulatory, and commercial obstacles, and enhancing the ability of systems to work naturally with care providers, patients, and patients' families.  The report also identifies capabilities for fluent interactions and effective partnering with people as key to achieving the promise of AI technologies for enhancing education. Major challenges to realizing the potential for AI to address the needs of low-resource communities include design of methods to cooperate with agencies and organizations working in those communities and the development of trust of AI technologies by these groups and by the communities they serve. Such challenges arise as well in the arenas of public safety and security. In the domain of employment and workplace, while noting that AI-capable systems will replace people in some kinds of jobs, the report predicts that AI capabilities are more likely to change jobs by replacing tasks than by eliminating jobs. It highlights the role of social and political decisions in approaching a range of societal challenges that will arise as work evolves in response to AI technologies, and it argues these challenges should be addressed now.

In assessing "What's Next?" in AI research, the report notes that "as it becomes a central force in society, the field of AI is shifting toward building intelligent systems that can collaborate effectively with people, and that are more generally human-aware". It identifies several currently "hot areas" of AI research and application. In the area of machine learning, it describes efforts in scaling to work with very large data sets, deep learning, and reinforcement learning. Robotics, computer vision, and natural language processing (including spoken language systems), already incorporated into a variety of applications, have made great strides recently and are poised to made further advances. Research in two relatively newer areas, collaborative systems and crowdsourcing/human computation, are developing methods, respectively, for AI systems to work effectively with people and for people to assist AI systems in computations which are more difficult for machines than for people.  Other research areas the report highlights are algorithmic game theory and computational social choice, the Internet of Things, and neuromorphic computing.

The report concludes with a section on policy and legal issues, which summarizes the Study Panel's views on the current state of regulatory statutes relevant to AI technologies and contains its recommendations to policy makers. It notes, importantly, that, "The measure of success for AI applications is the value they create for human lives. In that light, they should be designed to enable people to understand AI systems successfully, participate in their use, and build their trust." The report encourages "vigorous and informed debate" about AI capabilities and limitations, recommending that much broader and deeper understanding of AI is needed in government at all levels to enable expert assessments within government of AI technologies, programmatic objectives, and overall societal values. It argues that industry needs to formulate and deploy best practices as well as that AI systems should be open or amenable to reverse engineering so that they can be evaluated adequately with respect to such crucial issues as fairness, security, privacy, and social impacts by disinterested academics, government experts, and journalists. It also notes the importance of expertise across a wide variety of disciplinary areas being brought to bear on assessing societal impact and thus the need for increased public and private funding for interdisciplinary studies of the societal impacts of AI.

We list below several of the report's most important messages and findings, as reflected in callouts in the report. We hope this sample provides readers with a sense of the scope of the report and encourages them to read the report (at least at one of the levels of detail provided) to find out more.

- General Observations:
    - Like other technologies, AI has the potential to be used for good or nefarious purposes. A vigorous and informed debate about how to best steer AI in ways that enrich our lives and our society is an urgent and vital need. (p. 48)

    - As a society, we are underinvesting resources in research on the societal implications of AI technologies. Private and public dollars should be directed toward interdisciplinary teams capable of analyzing AI from multiple angles. (p. 44)

    - Misunderstandings about what AI is and is not could fuel opposition to technologies with the potential to benefit everyone. Poorly informed regulation that stifles innovation would be a tragic mistake. (p. 10)

- Potential Near-term Applications and Design Constraints:
    - While drawing on common research and technologies, AI systems are specialized to accomplish particular tasks. Each application requires years of focused research and a careful, unique construction. (p. 5)

    - AI-based applications could improve health outcomes and quality of life for millions of people in the coming years - but only if they gain the trust of doctors, nurses, and patients. (p. 26)

    - Though quality education will always require active engagement by human teachers, AI promises to enhance education at all levels, especially by providing personalization at scale. (p. 31)

    - With targeted incentives and funding priorities, AI technologies could help address the needs of low-resource communities. Budding efforts are promising. (p. 35)

- Societal concerns:
    - As highlighted in the movie *Minority Report* (authors note: and subsequently in *ProPublica* [1]), predictive policing tools raise the specter of innocent people being unjustifiably targeted. But well-deployed AI prediction tools have the potential to actually remove or reduce human bias. (p. 37)
    - AI will likely replace tasks rather than jobs in the near term, and will also create new kinds of jobs. But the new jobs that will emerge are harder to imagine in advance than the existing jobs that will likely be lost. (p. 38)
    - As AI applications engage in behavior that, were it done by a human, would

constitute a crime, courts and other legal actors will have to puzzle through whom to hold accountable and on what theory. (p. 46)

**Reactions and Uses**

Even more than when the AI100 project was first planned, we are at a crucial juncture in determining how to deploy AI-based technologies in ways that support societal needs and promote rather than hinder democratic values of freedom, equality, and transparency. The philosopher J. Moor has noted ([3], p. 267) that in ethical arguments, most often people agree on values but not on the facts of the matter. The AI100 report aims to bring AI expertise to the forefront so that the challenges as well as the promises of technologies that incorporate AI methods can be understood and properly assessed.

Although the report's impact over time remains to be seen, we hope it will set a strong precedent for future AI100 Study Panels. We have been gratified to have seen over the past years that the report seems to have succeeded in this aim, and the larger AI 100 goals, in several ways. Shortly after it was released, on September 1st, 2016, the report was covered widely in the press, including in the New York Times, the Christian Science Monitor, on NPR, the BBC, and CBC radio. It helped shape a series of OSTP sponsored workshops and the reports that emanated from them [2]. Requests for permission to translate the report into several languages have indicated worldwide interest. We have been asked to organize workshops for various governmental and scientific organizations and to give talks in many settings. The study panel chair (and co-author of this column) was invited to speak by the Prime Minister of Finland on the occasion of him announcing a new "AI Strategy" for Finland in February of 2017 (see http://valtioneuvosto.fi/live?v=/vnk/events-seminars/professori-peter-stonen-puhe-tekoalyseminaarissa). The report is also being used in AI classes in various ways.

**Looking Forward**

AI technologies are becoming ever more prevalent, and opinions on their impact on individuals and societies vary widely, from those the Study Panel considered overly optimistic to others it considered overly pessimistic. The need for the general public, government and industry to have reliable, balanced information is of increasing importance. The AI100 project aims to fill that need. This first report is an important initial step, launching the long-term project. It crucially illuminates the enormous technical differences between AI technologies that are developed and targeted towards specific application domains and a "general purpose AI" capability that can be incorporated into any device to make it more intelligent. The former is the focus of much research and business development, while the latter remains science fiction. It is quite tempting to think that if AI technologies can help drive your car, then they ought to also be able to fold your laundry, but these two activities make very different types of demands on reasoning. They require very different algorithms and capabilities. People do both and a full range of equally distinct activities requiring intelligence of various sorts. Current AI applications are based on specialized domain-specific methods, however, and the normal human inclination to generalize from one intelligent behavior to seemingly related ones leads people astray when assessing machine capabilities. This first AI100 report aims to provide insights to its readers enabling them to better assess the implications of any AI success for other open challenges, as well as to alert them to the societal and ethical issues that we must address together as AI pervades ever more areas of daily life.

Since publishing the report, the AI100 project has begun a complementary effort, the "AI Index". This ongoing tracking activity, led by a steering committee comprising Yoav Shoham, Ray

Perrault, Erik Brynjolfsson, Jack Clark, John Etchemendy, Terah Lyons, and James Maniyka, complements the major studies originally envisioned for AI100 by providing annual reports and, in the future, an ongoing blog that augment major AI100 studies occurring every five years. The AI Index will follow various facets of AI – including ones related to the volume of activity, technological progress, and societal impact – as determined by a broad advisory panel with advice also from the AI100 Standing Committee. Like the study panel reports, the Index aims to provide information on the status of AI that is useful for those outside the field and for those engaged in developing AI technologies as well as those actively involved in AI research and applications, for policy makers and business executives, and for the general public. This nascent effort issued its first report in December 2017.

The AI100 project (https://ai100.stanford.edu/) welcomes advice as it plans its next report, as does the AI Index (http://aiindex.org/). We look forward to following, and continuing to help shape, AI100's trajectory over the coming years.

References:

1. J. Angwin et al. "Machine Bias: There's software used across the country to predict future criminals. And it's biased against blacks." *ProPublica*, May 23 (2016).

2. Ed Felten and Terah Lyons, "The Administration's Report on the Future of Artificial Intelligence", https://obamawhitehouse.archives.gov/blog/2016/10/12/administrations-report-future-artificial-intelligence

3. J. H. Moor. "What is Computer Ethics?" *Metaphilosophy*, 16:4, 1985.

4. Peter Stone, Rodney Brooks, Erik Brynjolfsson, Ryan Calo, Oren Etzioni, Greg Hager, Julia Hirschberg, Shivaram Kalyanakrishnan, Ece Kamar, Sarit Kraus, Kevin Leyton-Brown, David Parkes, William Press, AnnaLee Saxenian, Julie Shah, Milind Tambe, and Astro Teller. "Artificial Intelligence and Life in 2030." One Hundred Year Study on Artificial Intelligence: Report of the 2015-2016 Study Panel, Stanford University, Stanford, CA, September 2016. Doc: http://ai100.stanford.edu/2016-report.

Barbara J. Grosz (grosz@eecs.harvard.edu) is Higgins Professor of Natural Sciences in the Computer Science faculty of the John A. Paulson School of Engineering and Applied Sciences at Harvard University, Cambridge, MA, USA and a member of the External Faculty of Santa Fe Institute, Santa Fe, NM, USA. She was Inaugural Chair of the Standing Committee for the One Hundred Year Study on Artificial Intelligence.

Peter Stone (pstone@cs.utexas.edu) is the David Bruton, Jr. Centennial Professor in the Department of Computer Science at The University of Texas at Austin, Austin, TX, USA. He was Chair of the Inaugural Study Panel of the One Hundred Year Study on Artificial Intelligence.